\documentclass[letterpaper, 10 pt, conference]{ieeeconf}  

\IEEEoverridecommandlockouts                              

\makeatletter
\let\NAT@parse\undefined
\makeatother

\usepackage[sort&compress,numbers,sectionbib]{natbib}

\usepackage{graphicx} 
\usepackage{amsmath} 
\usepackage{amssymb}  
\usepackage{algorithm}
\usepackage{xcolor}
\usepackage{algpseudocode}
\usepackage{subfig}
\usepackage{multirow}
\usepackage{booktabs}
\usepackage{hyperref}
\hypersetup{bookmarksopen,bookmarksnumbered,
pdfpagemode=UseOutlines,
colorlinks=true,
linkcolor=teal,
anchorcolor=teal,
citecolor=teal,
filecolor=teal,
menucolor=teal,
urlcolor=teal
}

\usepackage{balance}

\title{\LARGE \bf
Cook2LTL: Translating Cooking Recipes to LTL Formulae\\ using Large Language Models
}

\author{Angelos Mavrogiannis$^1$, Christoforos Mavrogiannis$^2$, and Yiannis Aloimonos$^1$
\thanks{$^1$Department of Computer Science,
        University of Maryland, College Park, 8125 Paint Branch Dr, College Park, MD 20742, USA.
        {\tt\small angelosm@cs.umd.edu, jyaloimo@cs.umd.edu}}%
\thanks{$^2$Department of Robotics, University of Michigan, Ann Arbor, MI, 48105. {\tt\small cmavro@umich.edu}.
}
\thanks{We release our \href{https://github.com/angmavrogiannis/Cook2LTL}{code} and a \href{https://youtu.be/5Q5cdcPN_2E}{video} with an example simulation rollout.}
}

\begin{document}

\maketitle
\thispagestyle{empty}
\pagestyle{empty}

\begin{abstract}


Cooking recipes are challenging to translate to robot plans as they feature rich linguistic complexity, temporally-extended interconnected tasks, and an almost infinite space of possible actions. Our key insight is that combining a source of cooking domain knowledge with a formalism that captures the temporal richness of cooking recipes could enable the extraction of unambiguous, robot-executable plans. 
In this work, we use Linear Temporal Logic (LTL) as a formal language expressive enough to model the temporal nature of cooking recipes. Leveraging a pretrained Large Language Model (LLM), we present Cook2LTL, a system that translates instruction steps from an arbitrary cooking recipe found on the internet to a set of LTL formulae, grounding high-level cooking actions to a set of primitive actions that are executable by a manipulator in a kitchen environment. Cook2LTL makes use of a caching scheme that dynamically builds a queryable action library at runtime. We instantiate Cook2LTL in a realistic simulation environment (AI2-THOR), and evaluate its performance across a series of cooking recipes. We demonstrate that our system significantly decreases LLM API calls ($-51\%$), latency ($-59\%$), and cost ($-42\%$) compared to a baseline that queries the LLM for every newly encountered action at runtime. 

\end{abstract}

\section{INTRODUCTION}

To be useful in household environments, robots may need to understand and execute instructions from novice users. Natural language is possibly the easiest way for users to provide instructions to robots but it is often too vague. This motivates the need for mapping natural language to actionable, robot-executable commands. This is a challenging problem, especially for complex activities that include temporally correlated subtasks, such as following instructions in a manual, or performing a delicate assembly task. 

\begin{figure}[t]
    \centering
    \includegraphics[width=\columnwidth]{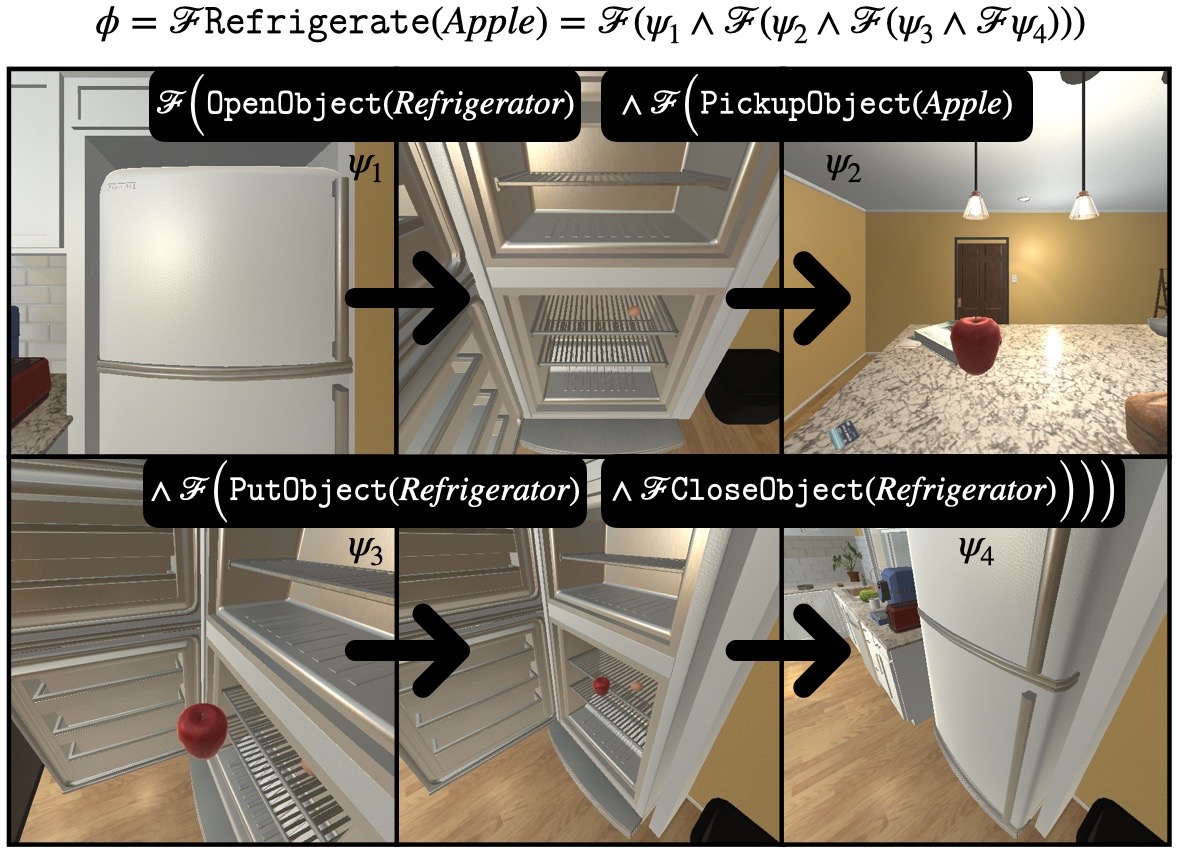}
    \caption{\textbf{Cook2LTL in AI2-THOR}~\citep{kolve2017ai2}: The robot is given the instruction \textit{Refrigerate the apple}. Cook2LTL produces an initial LTL formula $\phi$ (\textit{top left}); then it queries an LLM to retrieve the low-level admissible primitives for executing the action; finally it generates a formula consisting of 4 atomic propositions ($\psi_1,\psi_2,\psi_3,\psi_4$) that provide the required task specification and yield these consecutive scenes.}
    \label{fig:AI2-THOR_simulation}
\end{figure}

In this paper, we focus on translating cooking recipes into executable robot plans. Cooking is one of the most common household activities and poses a unique set of challenges to robots~\citep{bollini2013interpreting}. It usually requires following a recipe, written assuming that the reader has some background experience in cooking and commonsense reasoning to understand and complete the instruction steps. Recipes often feature ambiguous language~\citep{malmaud2014cooking}, 
such as omitting arguments that are easily inferred from context (the known ``Zero Anaphora'' problem~\citep{jiang2020recipe}; see Fig.~\ref{fig:annotation} where the direct object of the verb ``cook" is missing), 
or, more crucially, underspecified tasks 
under the assumption that the reader possesses the necessary knowledge to fill in the missing steps. For example, recipes with eggs do not explicitly state the prerequisite steps of cracking them and extracting their contents. Additionally, although inherently sequential, recipes often include additional explicit sequencing language (e.g. until, before, once) that clearly defines the temporal action boundaries.

Motivated by these observations, our key insight is that combining a source of cooking domain knowledge with a formalism that captures the temporal richness of cooking recipes could enable the extraction of unambiguous, robot-executable plans. 
Our \textbf{main contribution} is Cook2LTL, a system that receives a cooking recipe in natural language form, reduces high-level cooking actions to robot-executable primitive actions through the use of LLMs, and produces unambiguous task specifications written in the form of LTL formulae (See Fig.~\ref{fig:AI2-THOR_simulation}). These plans are then suitable for use in downstream robotic tasks. We build and evaluate our method based on a subset of recipes from the Recipe1M+ corpus~\citep{marin2019learning}. We run Cook2LTL on these recipes and show that by caching the action reduction policy, we incrementally build a queryable action library and limit proprietary LLM API calls with significant benefits in cost ($-42\%$) and computation time ($-59\%$) compared to a baseline that queries the LLM for every unseen action at runtime. We demonstrate the transferability of Cook2LTL to a robotic platform through experiments in a simulated kitchen in AI2-THOR~\citep{kolve2017ai2}.

\begin{figure*}[t]
    \centering
    \includegraphics[scale=0.2]{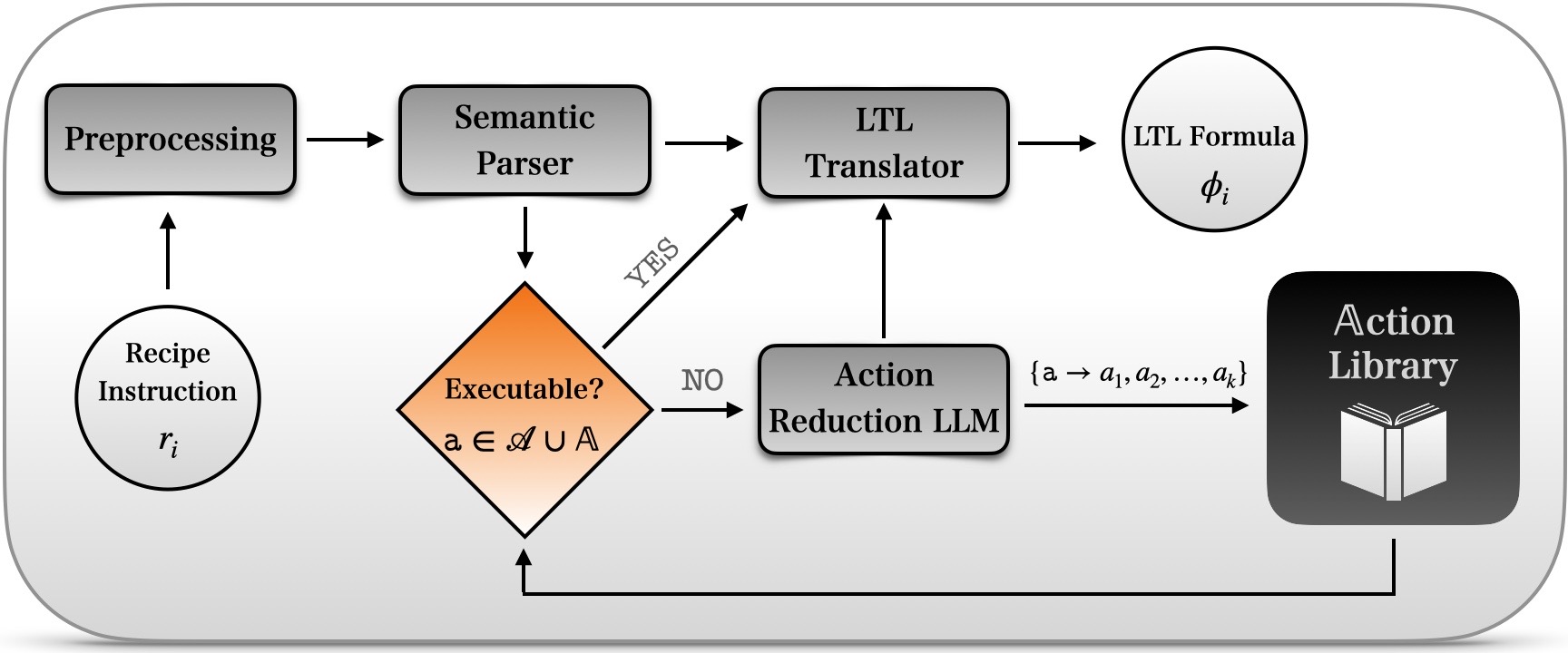}
    \caption{\textbf{Cook2LTL System:} The input instruction $r_i$ is first preprocessed and then passed to the semantic parser, which extracts meaningful chunks corresponding to the categories $\mathcal{C}$ and constructs a function representation $\mathtt{a}$ for each detected action. If $\mathtt{a}$ is part of the action library $\mathbb{A}$, then the LTL translator infers the final LTL formula $\phi$. Otherwise, the action is reduced to a sequence of lower-level admissible actions \{$a_1,a_2,\dots a_k\}$ from $\mathcal{A}$, and the reduction policy is cached to $\mathbb{A}$ for future use. The LTL translator then yields the final LTL formulae based on the derived actions.}
    \label{fig:pipeline}
\end{figure*}

\section{RELATED WORK}

\textbf{Robotic Cooking:} Cooking has been an important means of studying action understanding~\citep{poeticon,yang2015robot,pastra2012minimalist,beetz2011robotic}. The EU project POETICON~\citep{poeticon} viewed cognitive systems as a set of languages \{natural, visual, motoric\} and integrated these languages towards understanding cooking actions. Along these lines,~\citet{yang2015robot} processed YouTube videos using Convolutional Neural Networks (CNNs) and a grammatical approach~\citep{pastra2012minimalist} to produce parse trees that could be used for generating cooking actions. A few works have built end-to-end cooking systems that implement textual recipes on real robots~\citep{bollini2013interpreting,beetz2011robotic}, but are restricted to completing a roughly specific task (e.g., baking~\citep{bollini2013interpreting}, and making pancakes~\citep{beetz2011robotic}) and hence can only deal with a limited subset of recipes. On the other hand, more versatile commercial solutions (e.g., the Moley kitchen~\citep{moley}) are expensive and to the best of our knowledge cannot handle unseen recipes in real time.

Although our approach has not been applied on a real-world hardware platform, our AI2-THOR simulation~\citep{kolve2017ai2} in Sec.~\ref{sec:results} demonstrates its transferability to a real robot while allowing the system to adapt to new recipes.

\textbf{LLM planning:} Several works have grounded high-level actions to a well-defined set of actions for task planning using textual LLM-~\citep{pmlr-v205-ichter23a,huang2022inner,huang2022language,madaan2022language,vemprala2023chatgpt,singh2022progprompt,lin2023text2motion} or multimodal LLM-based~\citep{wu2023tidybot,zeng2022socratic,driess2023palm} approaches. Our interest lies in the former category given the unimodal nature of our text-to-robot action approach. These textual LLM-based works have shown great performance but come with certain limitations. For instance, the framework of~\citet{pmlr-v205-ichter23a} cannot handle open-vocabulary or combinatorial tasks, the one by~\citet{huang2022inner} might produce action plans including items that are not present in the current environment, and the model of~\citet{huang2022language} does not guarantee that the returned actions are admissible in the current context. Some of these works~\citep{madaan2022language,vemprala2023chatgpt,singh2022progprompt,wang2023demo2code} have leveraged programming language structures as an expressive tool for efficiently representing a rich set of task procedures in the LLM prompts. In the context of cooking, \citet{wang2023demo2code} have used LLMs to break down high-level cooking actions into actionable plans. However, their approach requires access to demonstrations of the intermediate steps of the cooking task at hand.

In our work, we adapt the methodology proposed by~\citet{singh2022progprompt}, where the task planning problem is formulated as a pythonic few-shot prompting scheme. The prompt consists of a pythonic import of a set of primitive actions, a definition of a list of available objects, and a few example task plans in the form of pythonic functions. Their experiments showed that prompting an LLM for task planning in a programmatic fashion outperforms verbose descriptive prompts by restricting the output plan to the constrained set of primitive actions and objects available in the current environment.


\textbf{Natural Language to LTL:} LTL was initially used in formal verification for computer programs~\citep{pnueli1977temporal}. Since then, it has been extensively used in robotics~\citep{fainekos2005temporal, kress2009temporal,smith2011optimal} as a formalism that enables the extraction of guarantees on robot performance given a robot model, a high-level description of its actions, and a class of admissible environments. There has been considerable work on translating natural language instructions to task specifications in the form of LTL~\citep{gopalan2018sequence,patel2020grounding,wang2021learning,berg2020grounding,liu2023lang2ltl,pan2023data} and its variants~\citep{mohammadinejad2022interactive,chen2023autotamp}. Most approaches try to address the main bottleneck which is the high cost of obtaining annotations of natural language with their equivalent LTL logical forms. \citet{gopalan2018sequence} orchestrate a data collection and augmentation pipeline to build a synthetic domain and translate natural language to LTL formulae using Seq2Seq models~\citep{bahdanau2014neural}. Alternatively, \citet{patel2020grounding,wang2021learning} learn from trajectories paired with natural language to reduce the need for human annotation, however a lot of trajectories are required to implicitly supervise the translator. \citet{berg2020grounding,liu2023lang2ltl} ground referring expressions to a known set of atomic propositions and translate to LTL formulae using Seq2Seq models~\citep{gu2016incorporating} and LLMs~\citep{brown2020language}, respectively. Similarly, \citet{pan2023data,chen2023nl2tl} use the paraphrasing abilities of LLMs to generate synthetic datasets tackling the scarcity of labeled LTL data.

Our approach is more similar to the work of~\citet{chen2023nl2tl} and~\citet{hsiung2022generalizing}, abstracting natural language to an intermediate representation layer before grounding to the final atomic propositions. An important limitation of these methods is that they are based on thoroughly curated datasets or well-structured synthetic data generation pipelines. On the contrary, we deal with unstructured free-form recipe text scraped from the internet. Moreover, most of these works in embodied settings have mainly been applied to navigation and simple pick-and-place tasks or combinations of these. Our web-scraped cooking recipe corpus offers a richer and more diverse action space. 

\section{PRELIMINARIES}
\label{sec:prelim}
This section provides a short background on LTL and LLMs, which are the tools we are using in our pipeline.
\subsection{Linear Temporal Logic}
LTL is a temporal logic that was developed for formal verification of computer programs through model checking~\citep{pnueli1977temporal}. It is suitable for expressing task specifications and verifying system performance in safety-critical applications. These task specifications are expressed through the use of this grammar:
\begin{equation} \label{eq:LTL_def} 
    \phi::=p\,|\,\lnot p\,|\,\phi_1\wedge\phi_2\,|\,\phi_1\vee\phi_2\,|\,\mathcal{G}\,\phi\,|\,\mathcal{F}\,\phi\,|\,\phi_1\,\mathcal{U}\,\phi_2
\end{equation}
where $\phi$ is a task specification, $\phi_1$ and $\phi_2$ are LTL formulae, and $p\in$ is an atomic proposition drawn from a set $\mathcal{P}$ of atomic propositions (APs). $\lnot,\wedge,\vee$ are the known symbols from standard propositional logic denoting negation, conjunction, and disjunction, respectively. As an extension, LTL supports additional temporal operators. More specifically, $\mathcal{G}\,\phi$ denotes that $\phi$ holds globally, $\mathcal{F}\,\phi$ denotes that $\phi$ must eventually hold, and $\phi_1\,\mathcal{U}\,\phi_2$ indicates that $\phi_1$ must hold for all time steps until $\phi_2$ becomes true for the first time. In this work, we utilize LTL as a formal language to express temporally-extended cooking tasks.

\subsection{Large Language Models}
Given a piece of text $W=\{w_1,w_2,\dots,w_n\}$ consisting of $n$ words $w_i, i = 1,\dots, n$, a language model estimates the probability $p(W)$. This is done in an auto-regressive manner, leveraging the chain rule to factorize the probability~\citep{bengio2000neural}:
\begin{equation} \label{eq:LLM_def}
p(W)=p(w_1,w_2,\dots,w_n)=\prod_{i=1}^{n}p(w_{i}|w_1,\dots,w_{i-1})
\end{equation}
Generating text can then be achieved recursively. Given a set of preceding words $\{w_1,w_2,\dots,w_{i-1}\}$, the model estimates the probability distribution for the next word $p(w_i|w_1,\dots,w_{i-1})$. LLMs, such as BERT~\citep{devlin2018bert} and GPT-3~\citep{brown2020language} are pre-trained on large-scale internet corpora and have dominated across a series of downstream natural language processing (NLP) tasks~\citep{srivastava2022beyond}. In this work, we leverage the domain knowledge encoded into such models in order to reduce high-level tasks to actions on a lower level of abstraction.

\section{TRANSLATING COOKING RECIPES TO LTL FORMULAE}

\label{sec:translation}
\subsection{Problem Statement}
\label{sec:prob_statement}

Consider a robot in a kitchen, equipped with a limited set of primitive actions $\mathcal{A}$. 
We assume that a primitive action in a cooking environment can be described by a set of salient categories $\mathcal{C}=$\texttt{\{Verb, What?, Where?, How?, Time, Temperature\}}. We define an action description $\mathtt{a}$ as a function consisting of a main \texttt{Verb} as the function name, with a set of one or more of the other categories as its parameters:
\begin{equation*}
    \mathtt{a}=\mathtt{Verb(What?,Where?,How?,Time,Temperature)}
\end{equation*}
The robot is tasked with executing a cooking recipe $R$ that consists of a list of $k$ instruction steps $\{r_1,r_2,\dots,r_k\}$, where each instruction step $r_i$ is an imperative sentence in natural language describing a robot command. Each instruction step $r_i$ may include one or more cooking actions. Our goal is to generate a set of task specifications written in the form of a set of LTL formulae $\Phi=\{\phi_1,\phi_2,\dots,\phi_n\}$ that implement the recipe under the constraint of \textbf{only} including actions that belong to the set of primitive actions $\mathcal{A}$ that the robot is capable of executing.

\subsection{System Architecture}

To solve this problem, we propose Cook2LTL, the system architecture summarized in Fig.~\ref{fig:pipeline}. Given an instruction $r_i$ and a set of actions $\mathcal{A}$, Cook2LTL:


\begin{enumerate}
    \item Semantically parses $r_i$ into a function representation $\mathtt{a}$ for every detected high-level action.
    \item Reduces each high-level action $\mathtt{a}\notin\mathcal{A}$ to a combination of primitive actions from $\mathcal{A}$.
    \item Caches the action reduction policy for future use, thereby gradually building an action library that consists of parametric functions that express high-level cooking actions in the form of primitive actions.
    \item Translates $r_i$ into an LTL formula $\phi_i$ with function representations as atomic propositions.
\end{enumerate}
Algorithmically, these steps are summarized in Alg.~\ref{alg:Cook2LTL}. In the following subsections, we expand on the components of Cook2LTL in more detail.




\begin{algorithm}
\caption{Cook2LTL}\label{alg:Cook2LTL}
\textbf{Input:} A high-level instruction step $r$, a set of primitive actions $\mathcal{A}$, and an action library $\mathbb{A}$\\
\textbf{Output:} An LTL action formula $\phi$
\begin{algorithmic}[1]
\State $\mathbb{A}\gets\mathbb{A}\cup\mathcal{A}$
\State $r\gets f_{PRE}(r)$ \Comment{Preprocessing}
\State $\{\mathtt{a_1},\mathtt{a_2},\dots,\mathtt{a_n}\} \gets f_{SP}(r)$ \Comment{Semantic Parsing}
\State $A \gets\{\mathtt{a_1},\mathtt{a_2},\dots,\mathtt{a_n}\}$
\State $\phi \gets f_{LTL}(\mathtt{a_1},\mathtt{a_2},\dots,\mathtt{a_n})$ \Comment{Initial LTL Translation}
\For{$\mathtt{a}_i\in A$}                    
    \If{$\mathtt{a_i} \notin \mathbb{A}$}
        \State $\{a_1,a_2,\dots,a_k\} \gets f_{AR}(\mathtt{a}_i)$ \Comment{Action Reduction}
        \State $\mathtt{a_i} \gets \{a_1,a_2,\dots,a_k\}$
        \State $\mathbb{A} \gets \mathbb{A}\cup\{\mathtt{a}\rightarrow a_1,a_2,\dots,a_k\}$ \Comment{Caching}
    \EndIf
\EndFor
\State $\phi \gets f_{LTL}(A)$ \Comment{Final LTL Translation}
\State \Return $\phi$
\end{algorithmic}
\end{algorithm}

\subsection{Semantic Parsing and Data Annotation}
\label{sec:data_annotation}
Our translation system requires a semantic parsing module capable of extracting meaningful chunks corresponding to the parametric function representation components of a cooking action. To this end, we fine-tune a named entity recognizer with the addition of salient categories $\mathcal{C}$ as labels. We choose a neural approach over a syntactic parse because the latter would require arduous manual rule crafting for every different mapping of part-of-speech (POS) tags to these categories. Additionally, explicit POS-tagging-based approaches often struggle with handling the intricacies of cooking discourse, such as imperative form sentences omitting context-implicit parts of speech.

In the absence of a labeled dataset with a schema matching $\mathcal{C}$, we create our own data building upon the large cooking recipe dataset Recipe1M+~\citep{marin2019learning}. Specifically, we consider a subset of $100$ recipes from Recipe1M+, leading to $1000$ recipe instruction steps. We use brat~\citep{brat} to manually annotate chunks in each step corresponding to the following salient categories: $\mathcal{C}=$\texttt{\{Verb, What?, Where?, How?, Temperature, Time\}}, which is a similar annotation scheme as the one seen in recent work~\citep{papadopoulos2022learning}.

\begin{figure} 
    \subfloat[Salient categories $\mathcal{C}$ considered for semantic parsing.]{\label{fig:categories}\includegraphics[width=0.9\columnwidth]{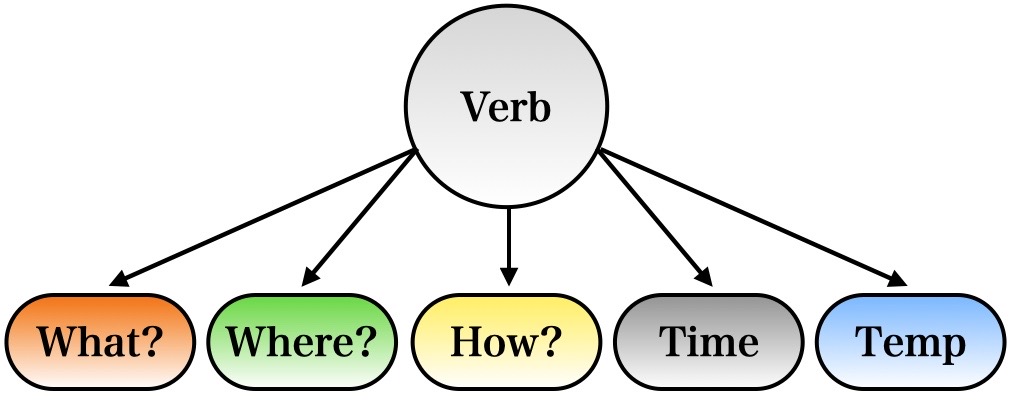}}
    
    \subfloat[Recipe steps annotated with the salient categories $\mathcal{C}$]{\label{fig:annotation}\includegraphics[width=0.9\columnwidth]{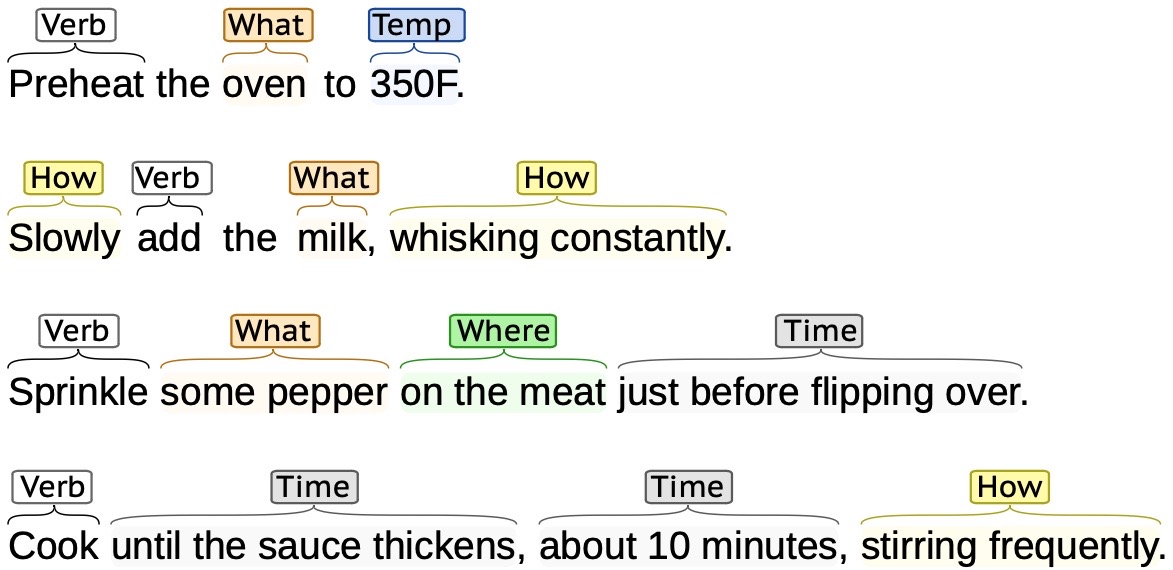}}
    \caption{We annotate Recipe1M+~\citep{marin2019learning} instruction steps with the salient categories $\mathcal{C}=$\texttt{\{Verb, What?, Where?, How?, Temperature, Time\}} and fine-tune a named entity recognizer to segment chunks corresponding to $\mathcal{C}$.}
    \label{fig:categories_annotation}
\end{figure}

Fig.~\ref{fig:categories_annotation} shows these categories and a set of example recipe steps taken from Recipe1M+~\citep{marin2019learning}. \texttt{Verb} is the main action verb in a recipe step. \texttt{What?} represents the direct object of the \texttt{Verb} and is often an ingredient, but can correspond to other entities such as a kitchen utensil or an appliance. \texttt{Where?} is usually a prepositional phrase, it implies a physical location (e.g. table, bowl) but can often be an ingredient to which the \texttt{Verb} applies. \texttt{How?} is usually either a gerund form of a verb, expressing concurrency and hence giving rise to a secondary cooking action, or complements the main cooking action (e.g. ``Drizzle \textit{with olive oil}"). The \texttt{Time} category consists of temporal expressions composed of keywords that are important for the translation of the commands to LTL formulae (\textit{until}, \textit{before} etc.). Finally, \texttt{Temperature} can explicitly list the degrees (e.g. 350F) to which food should be cooked or refer to a temperature-related state of some ingredient (e.g. \textit{medium heat}). These salient categories form the function representation of an action found in $r_i$.

\begin{figure*} 
    \centering
    \includegraphics[scale=0.3]{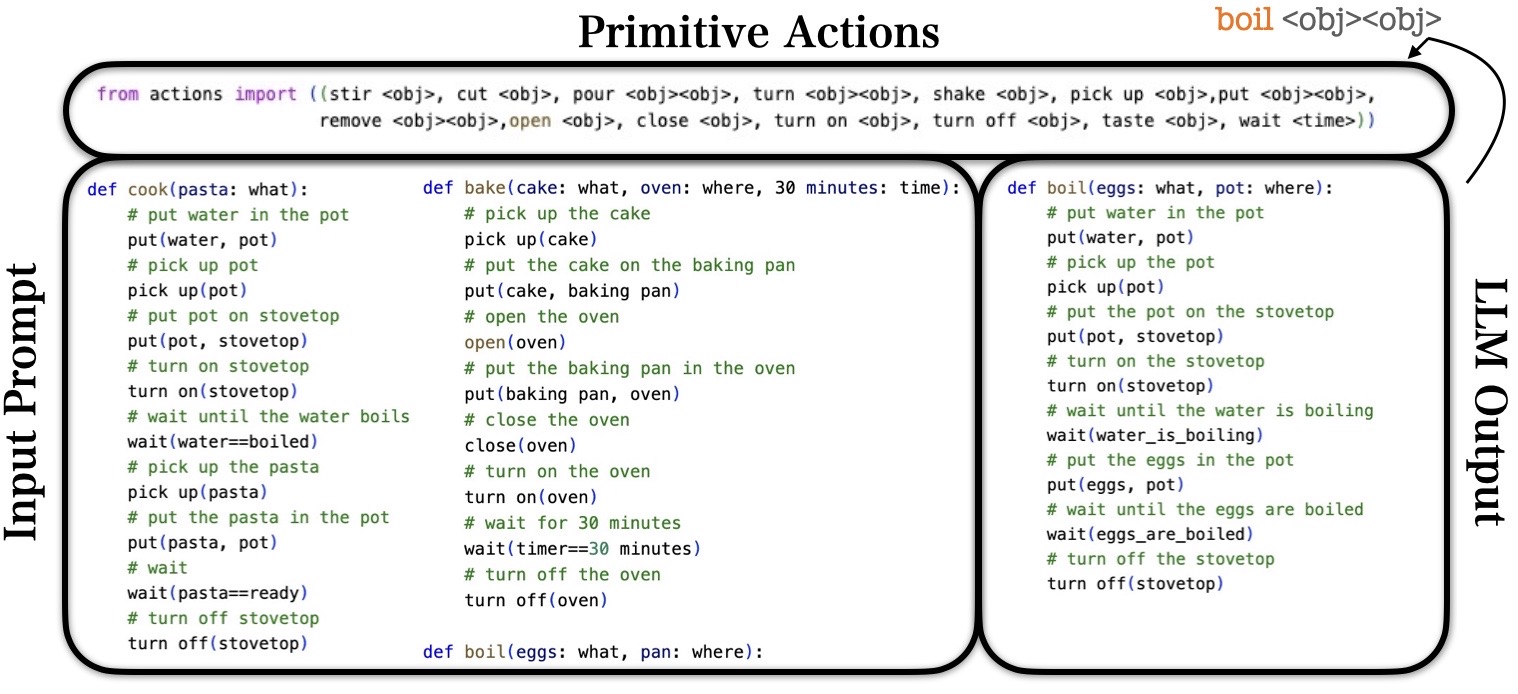}
    \caption{Inspired by ProgPrompt~\citep{singh2022progprompt}, Cook2LTL uses an LLM prompting scheme to reduce a high-level cooking action (e.g. \texttt{boil eggs}) to a series of primitive manipulation actions. The prompt consists of an import statement of the primitive action set and example function definitions of similar cooking tasks. The key benefit of using this paradigm is that it constrains the output action plan of the LLM to only include subsets of the available primitive actions. We extend this prompting scheme by reusing derived LLM policies. In this case, the action \texttt{boil} is added to future import statements in the input prompt, enabling the model to invoke the derived \texttt{boil} function which is now considered given to the system.}
    \label{fig:prompt_action_reduction}
\end{figure*}


\subsection{Reduction to Primitive Actions}

Some of the function representations captured in the previous step contain high-level actions that might not be directly executable by the robot, which can only execute actions that belong to the primitive set $\mathcal{A}$. Therefore, our system requires a module capable of mapping an action $\mathtt{a}\notin\mathcal{A}$ to an action $a\in\mathcal{A}$, if possible, or reducing $\mathtt{a}$ to a sequence of actions $a_1,a_2,\dots,a_k$ where $a_i\in\mathcal{A},i=1,2,\dots,k$. Our system initially checks whether $\mathtt{a}\in\mathcal{A}$ to validate a formula for execution, and if $\mathtt{a}\in\mathcal{A}$, $\mathtt{a}$ is forwarded to the LTL translator.


\textbf{LLM Action Reduction}: If $\mathtt{a}\notin\mathcal{A}$ we employ an LLM-based methodology inspired by the work in~\citep{singh2022progprompt} to extract a lower-level plan exclusively consisting of primitive actions from $\mathcal{A}$. Specifically, we design an input prompt consisting of: i) a pythonic import of the available actions in the environment, ii) two example function definitions decomposing high-level cooking actions into primitive sets of actions from $\mathcal{A}$, iii) the function representation $\mathtt{a}$ extracted by the semantic parsing module in the form of a pythonic function name with its parameters. As shown in~\citep{singh2022progprompt} and Fig.~\ref{fig:prompt_action_reduction}, the LLM follows the style and pattern of the input function and only includes available actions in the output. The key advantage of this method is the flexibility in changing the admissible primitive actions depending on the robot capabilities and the environment. This change can simply be achieved by modifying the primitive actions in the pythonic import.

\textbf{Action Library}: Extending ProgPrompt~\citep{singh2022progprompt}, every time we query the LLM for action reduction, we cache $\mathtt{a}$ and its action decomposition for future use through a dictionary lookup manner. This gradually builds a dynamic knowledge base in the form of an executable action library $\mathbb{A}$ consisting of various high-level actions along with their function bodies made out of primitive actions from $\mathcal{A}$. At runtime, instead of only checking whether a detected action $\mathtt{a}$ matches an action $a\in\mathcal{A}$, we additionally check if $\mathtt{a}\in\mathbb{A}$. In case there is a match, we replace $\mathtt{a}$ with the action in $\mathbb{A}$. Additionally, we add $\mathtt{a}$ to the pythonic import part of the prompt, allowing the model to invoke it when generating future policies (e.g. the action \texttt{boil} in Fig.~\ref{fig:prompt_action_reduction}). The key benefit comes from avoiding to continuously query an LLM for action reduction, thus replacing potential latency resulting from an LLM API call with a fixed $\mathcal{O}(1)$ dictionary lookup time. It also reduces the cost associated with querying a proprietary LLM API.

\subsection{LTL Translation}
The final step in our pipeline translates the intermediate function representations acquired from semantic parsing and action reduction into an LTL formula. The implicit sequencing of recipes is elegantly captured by the sequenced visit specification pattern~\citep{menghi2019specification}:
\begin{equation} \label{eq:LTL_seq_visit}
    F(l_1\wedge \mathcal{F}(l_2\wedge\dots\mathcal{F}l_n)))
\end{equation}
This pattern has been used~\citep{patel2019learning,patel2020grounding,liu2023lang2ltl} to model a visit of a set of locations $L=\{l_1,l_2,\dots,l_n\}$ in sequence one after the other in a navigational setting, adapted to the execution of consecutive cooking actions $\mathtt{a}_1,\mathtt{a}_2,\dots,\mathtt{a}_n$ in our case. 
Building on this pattern, we acquire conjunction, disjunction, and negation constituents for each segmented chunk corresponding to the categories $\mathcal{C}$ through a dependency parse. Then, we write down a formula $\phi$ which includes high-level actions $\mathtt{a}$ with a combination of the following LTL operators \{($\mathcal{F}$ : \texttt{Finally}), ($\wedge$ : \texttt{and}), ($\vee$ : \texttt{or}), ($\lnot$ : \texttt{not})\}. 
Every action $\mathtt{a}_i$ is translated to one or more primitive actions from $\mathcal{A}$. In the latter case, the generated low-level plan for $\mathtt{a}_i$ is parsed into a subformula $\psi_i$ based on Equation~\ref{eq:LTL_seq_visit}. The \texttt{Time} parameter passed to the action reduction LLM often includes explicit sequencing language (such as \textit{until}, \textit{before}, or \textit{once}). The LLM has been prompted to return a \texttt{Wait} function in these cases (see example in Fig.~\ref{fig:prompt_action_reduction}), which is then parsed into the ($\mathcal{U}$ : \texttt{until}) operator and substituted in $\psi$. The final formula $\phi$ consists of subformulae $\psi_1,\psi_2,\dots,\psi_n$ comprised by primitive actions in $\mathcal{A}$:
\begin{equation}
    \phi=F(\mathtt{a}_1\wedge \mathcal{F}(\mathtt{a}_2\wedge\dots\mathcal{F}\mathtt{a}_n)))= F(\psi_1\wedge \mathcal{F}(\psi_2\wedge\dots\mathcal{F}\psi_n)))
\end{equation} where:
\begin{equation}
    \begin{cases}
        \psi_i=a_i&,a_i\in\mathcal{A}\text{   ,or}\\
        \psi_i=f(a_1,a_2,\dots,a_k,\mathbb{O})&,\mathbb{O}=\{\mathcal{F},\wedge,\vee,\lnot,\mathcal{U}\}
    \end{cases}
\end{equation}



\section{EVALUATION}
\label{sec:evaluation}

\subsection{Ablation Study}
To investigate the performance of Cook2LTL, we conduct an ablation study against two variants. For each run, the input is a recipe from a held-out subset of Recipe1M+ and the output is a series of task specifications in the form of LTL formulae $\Phi$ towards executing the recipe under the constraints of admissible actions $\mathcal{A}$. In all the experiments we use the OpenAI API and the \textit{gpt-3.5-turbo} model. The initial preprocessing step consists of filling in the implicit objects (zero anaphora resolution) in the recipes and segmenting each recipe into sentences. We begin by deploying a partial version of our system (AR*) as a baseline, consisting of the preprocessing, semantic parsing, and action reduction modules. We expect that our action reduction policy adheres to the admissible actions of the environment by a significant amount. We incrementally add the functionality of invoking cached policies, first when encountering a primitive action (AR), and then when an action is found in the action library (AR+$\mathbb{A}$), starting from an empty library and gradually building it with the LLM-generated policies along the way. We anticipate a significant benefit in terms of computational load and cost efficiency resulting from capitalizing on reusable policies, compared to querying the action reduction LLM for every unseen action encountered at runtime. We formalized these insights into the following hypotheses:

\textbf{H1}: Our action reduction policy generation constrains the LLM output to the admissible actions $\mathcal{A}$ in our environment.

\textbf{H2}: Our enhanced Cook2LTL system that includes the action library component is more time- and cost-efficient than the baseline action reduction-comprised partial system.


To evaluate these hypotheses, our metrics are: 1. \textit{Executability (\%)}, which is the fraction of actions in the generated plan that are admissible in the environment; 2. \textit{Time (min} or \textit{sec)} which measures the runtime influenced by the LLM API calls; 3. \textit{Cost (\$)} which is the overall cost for a batch of experiments and depends on the number of input and output tokens; 4. the number of the LLM \textit{API calls}.

\begin{figure}
    \centering
    \includegraphics[width=\linewidth]{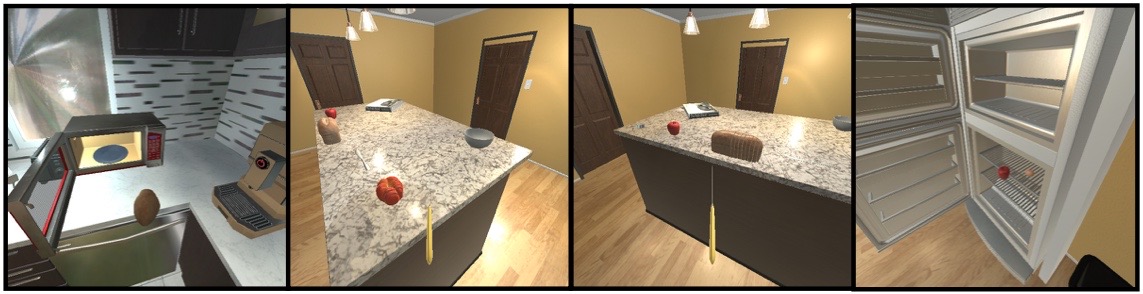}
    \caption{Tasks we tested Cook2LTL in AI2-THOR (left to right): \texttt{microwave the potato}; \texttt{chop the tomato}; \texttt{cut the bread}; \texttt{refrigerate the apple}.}
    \label{fig:screenshots}
\end{figure}

\subsection{Results \& Discussion}
\label{sec:results}

\begin{table}
    \centering
    \resizebox{\columnwidth}{!}{%
    \begin{tabular}{l|l|l|l}
        \toprule
        & \multicolumn{3}{c}{\textbf{Active Modules}}\\
        \hline
         \textbf{Metric} & AR* & AR & Cook2LTL (AR+$\mathbb{A}$) \\
        \hline
        Executability (\%) & $0.91\pm0.01$ & $0.92\pm0.01$ & $\mathbf{0.94\pm0.01}$ \\
        Time (min) & $14.85\pm1.05$ & $9.89\pm0.46$ & $\mathbf{6.05\pm0.12}$ \\
        Cost (\$) & $0.19\pm0.01$ & $0.16\pm0.00$ & $\mathbf{0.11\pm0.00}$ \\
        API calls (\#) & $275\pm0.00$ & $231\pm0.00$ & $\mathbf{134\pm0.00}$\\
        \bottomrule
    \end{tabular}
    }
\caption{Performance of Cook2LTL against baselines across $50$ Recipe1M+~\citep{marin2019learning} recipes (10 runs per recipe).}
\label{tab: ablation_recipes}
\end{table}

\begin{table}
    \centering
    \resizebox{\columnwidth}{!}{%
    \begin{tabular}{l|l|l|l|l}
        \toprule
        & \multicolumn{2}{c}{AR} & \multicolumn{2}{|c}{Cook2LTL (AR+$\mathbb{A}$)}\\
        \hline
         Task & SR (\%) & Time (sec) & SR (\%) & Time (sec) \\
        \hline
        Microwave the potato & $5.4\pm1.95$ & $27.29\pm3.66$ & $\mathbf{8\pm4.47}$ & $\mathbf{3.26\pm1.30}$\\
        Chop the tomato & $2.4\pm1.52$ & $16\pm0.96$ & $\mathbf{4\pm5.47}$ & $\mathbf{1.61\pm0.76}$\\
        Cut the bread& $\mathbf{9\pm0.71}$ & $12.85\pm0.84$ & $8\pm4.47$ & $\mathbf{1.12\pm0.16}$\\
        Refrigerate the apple & $7.6\pm0.55$ & $14.6\pm0.38$ & $\mathbf{8\pm4.47}$ & $\mathbf{1.56\pm0.44}$\\
        \bottomrule
    \end{tabular}
    }
\caption{We demonstrate the performance of Cook2LTL on $4$ simple cooking tasks in AI2-THOR. We observe that Cook2LTL (AR+$\mathbb{A}$) is time efficient but propagates initial incorrect LLM-generated sets of actions to subsequent runs.}
\label{tab: ai2thor_results}
\end{table}


Based on the quantitative results in Table~\ref{tab: ablation_recipes} we make the following observations regarding our hypotheses.

\textbf{H1}: Our first hypothesis is confirmed. In every part of the ablation study the system has a high executability with a maximum value of $94\%$ when using the action library. This is a natural consequence of incorporating a new action in the prompt every time it is decomposed to sub-actions by the LLM. The policies for the cached actions are now part of the system, and hence they are considered admissible in the environment, leading to an increased executability value.

\textbf{H2}: The enhanced action library-based Cook2LTL system (AR+$\mathbb{A}$) outperforms the baseline (AR*) and primitive action-focused variant (AR) in all 4 metrics. We have discovered that learning new action policies through prompting an LLM and reusing them in a dictionary lookup manner in subsequent recipes decreases the number of API calls by $51\%$ and $50\%$ compared to the AR* and AR versions of the system. Consequently, a lower number of API calls leads to a significantly reduced runtime and cost. More specifically, the integration of the action library into our system decreases runtime by $59\%$ and $42\%$ compared to the AR* and AR versions, and cost by $42\%$ and $31\%$, respectively.

\subsection{Demonstration in AI2-THOR}
We demonstrate the performance of Cook2LTL in a simulated AI2-THOR~\citep{kolve2017ai2} kitchen environment (See Fig.~\ref{fig:AI2-THOR_simulation}). AI2-THOR has a small set of ingredients and objects and hence cannot support the full execution of recipes found on the web; however the limited action space aligns with the notion of primitive actions and offers room for highlighting the key ideas of our system. To showcase the potential of our approach, we constructed a set of $4$ kitchen tasks that are admissible in AI2-THOR and executed them by invoking Cook2LTL. We assume that the kitchen is \textit{mise en place} so the locations of the objects are known to the agent. 
In AI2-THOR, we design a minimal parser that receives an LTL formula and converts it to a series of actions. We adapt the imported primitive actions and example functions in the prompt to the ones that are supported in the simulation. Fig.~\ref{fig:screenshots} contains screenshots from our experiments. We run $5$ sets of experiments where we execute each task $10$ consecutive times. We measure the success rate SR and execution time due to the LLM API calls and compare the performance of the AR and Cook2LTL (AR+$\mathbb{A}$) variants. The success rate is the fraction of executions that achieved the task-dependent goal conditions (e.g. \textit{tomato=sliced}) that we defined a priori. During our simulations we observe that Cook2LTL is still significantly more time efficient compared to baselines, however its SR is entirely dependent on the first LLM-generated plan, and fails when this plan is not executable (See Table~\ref{tab: ai2thor_results}).

\section{Limitations \& Future Work}
\label{sec:limitations_future_work}
\textbf{System}: We annotated a small part of the Recipe1M+ dataset~\citep{marin2019learning} with our salient categories but we would need more data to improve the entity recognizer for reliably transferring the system to a real-world robot. 
Finally, some actions being substituted by action library policies lead to non-executable plans. Our system would benefit from an additional mechanism that robustly ensures the correctness of the LLM-generated plans based on environment feedback.



\textbf{Sim2real}: AI2-THOR is not tailored towards simulating cooking tasks but rather supports the general area of task planning. Thus, we would need a cooking-specific simulator to support a more diverse set of recipes that correspond to the rich web-scraped recipes that we built our system on. In terms of transferring simulation to a real robot, we plan to use the Yale-CMU-Berkeley (YCB) Object and Model set~\citep{ycb} towards supporting a basic set of simple cooking tasks for benchmarking preliminary experiments.

\textbf{Task representation}: The final layer of our system uses LTL as an expressible notation tool capturing temporal task interdependence, but our system is compatible with other task representations, such as PDDL~\citep{mcdermott1998pddl}, which incorporates action preconditions and postconditions in the problem setting and has recently been explored with LLMs~\citep{silver2023generalized, liu2023llmp}.









\footnotesize{
\balance
\bibliographystyle{abbrvnat}
\bibliography{refs}
}

\end{document}